\documentclass[10pt,twocolumn,letterpaper]{article}

\usepackage{cvpr}
\usepackage{float}

\definecolor{cvprblue}{rgb}{0.21,0.49,0.74}
\usepackage[pagebackref,breaklinks,colorlinks,allcolors=cvprblue]{hyperref}

\def\paperID{5466} 
\def\confName{CVPR}
\def\confYear{2025}

\title{Shining Yourself: High-Fidelity Ornaments Virtual Try-on with Diffusion Model}

\author{
Yingmao Miao$^{1,2}$\footnotemark[1], Zhanpeng Huang$^2$\footnotemark[2], Rui Han$^2$, Zibin Wang$^2$, Chenhao Lin$^1$\footnotemark[2], Chao Shen$^1$\\
$^1$Xi'an Jiaotong University,\ $^2$ SenseTime Research\\
{\tt\small mym2017@stu.xjtu.edu.cn,} {\tt\small \{linchenhao, chaoshen\}@xjtu.edu.cn,}\\ {\tt\small \{huangzhanpeng, hanrui, wangzibin\}@sensetime.com}
}

\begin{document}
\twocolumn[{
\maketitle
\begin{figure}[H]
\hsize=\textwidth
\centering
\includegraphics[width=\textwidth]{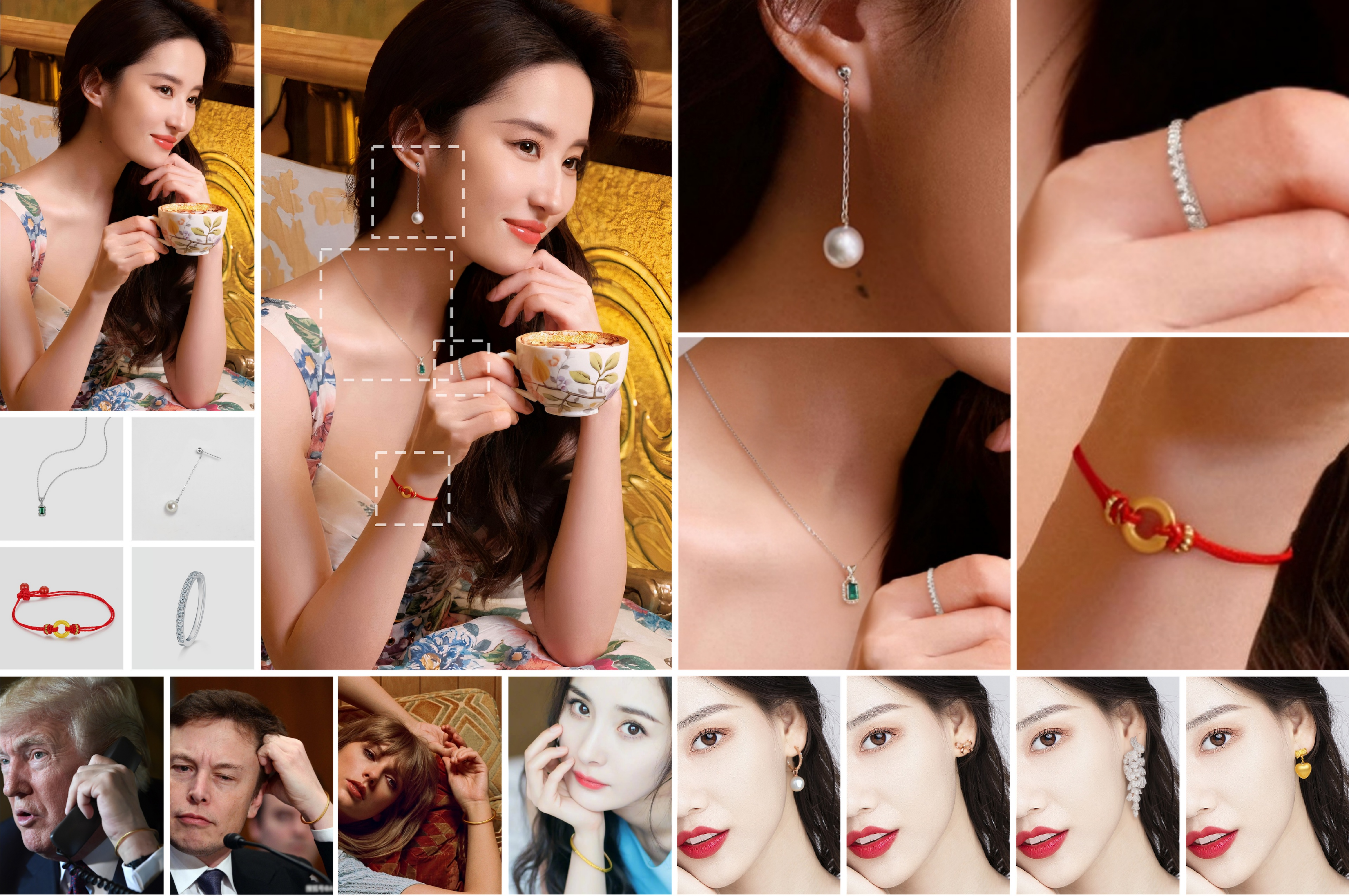}
\caption{\textbf{Shining Yourself.} We propose the virtual try-on task for ornaments including bracelets, rings, earrings, and necklaces for the first time. Our method achieves realistic virtual try-on results and high-fidelity identity preservation of ornament using pose-aware mask prediction and mask-guided attention. Project Page: https://shiningyourself.github.io/}
\label{first}
\end{figure}
}]
\footnotetext[1]{Completed during the internship at SenseTime Research}
\footnotetext[2]{Corresponding authors}

\begin{abstract}

While virtual try-on for clothes and shoes with diffusion models has gained attraction, virtual try-on for ornaments, such as bracelets, rings, earrings, and necklaces, remains largely unexplored. Due to the intricate tiny patterns and repeated geometric sub-structures in most ornaments, it is much more difficult to guarantee identity and appearance consistency under large pose and scale variances between ornaments and models. This paper proposes the task of virtual try-on for ornaments and presents a method to improve the geometric and appearance preservation of ornament virtual try-ons. Specifically, we estimate an accurate wearing mask to improve the alignments between ornaments and models in an iterative scheme alongside the denoising process. To preserve structure details, we further regularize attention layers to map the reference ornament mask to the wearing mask in an implicit way. Experimental results demonstrate that our method successfully wears ornaments from reference images onto target models, handling substantial differences in scale and pose while preserving identity and achieving realistic visual effects. 
\end{abstract}    
\section{Introduction}
\label{sec:intro}


As diffusion model~\cite{ddpm,ddim,beat,ldm,photo,ControlNet,hu2024animate} becomes the de facto standard in image generation, it's also widely adopted in the field of virtual try-on~\cite{zhu2023tryondiffusion,kim2024stableviton,xu2024ootdiffusion,yang2024texture,idm,zeng2024cat}. Given a reference image of an item 
and a target image of a model, the task is to get a preview of the fitting effect using image generation methods. Since no in-person wearing or physical fitting room is required, it has great potential for massive advertising materials generation in various applications of retail, e-commerce, and advertisement.


The main challenge of virtual try-on is how to generate a realistic fitting effect while preserving the fidelity of the garment. Massive efforts~\cite{kim2024stableviton,xu2024ootdiffusion,yang2024texture,idm,zeng2024cat} have been dedicated to solving the problem in the field of garments. The methods usually employ a network module (e.g., CLIP image encoder~\cite{clip} or ReferenceNet~\cite{animatediff,hu2024animate}) to extract garment features, which are injected into the process of diffusion denoising to preserve the identity and details of the garment. It has made great progress to be widely adopted for commercial use.

However, few works focus on virtual try-ons of ornaments such as bracelets, rings, earrings, and necklaces, despite significant practical demand. Virtual try-on for ornaments presents unique challenges that existing methods struggle to address effectively. 1) Most ornaments often feature intricate, small-scale geometric structures, such as rings and holes, that are difficult to preserve in virtual try-ons. In contrast, garments typically have sparser and/or repeated textures and lack complex geometric details. 2) For garment try-ons, generative visual artifacts can blend with natural cloth deformations and wrinkles, reducing their visibility. However, most ornaments are rigid or consist of rigid components, making any distortion or artifacts immediately noticeable. 3) Most garment virtual try-on methods require silhouette, skeleton, and semantic maps as additional inputs, while even the coarse silhouette mask of the ornament is not easy to depict due to its diverse structures and pose-depended occlusion.  
To tackle the problems, we propose a method to predict an accurate wearing mask to align the poses and scales between ornaments and models without additional inputs such as the silhouette, skeleton, or semantic maps. Our method also preserves structures and detailed features using a mask-guided attention of ornaments and models to preserve geometric structures. Specifically, to obtain a precise pose-aware wearing mask without explicit extraction of various maps of ornaments and models, we refine the mask from an input mask as coarse as a bounding box. A refined mask is then estimated using the intermediate features in the generation, which is further used as input to predict a more accurate one iteratively. As the mask indicates structure information, especially for multiple and tiny geometric structures, we formulate the attention layers to implicitly learn a mapping between the reference mask and the ground truth mask to preserve the structure patterns in ornaments. In summary, our contributions are as follows: 



\begin{itemize}
    \item To the best of our knowledge, we are the first to use diffusion models for virtual try-ons of various ornaments including bracelets, rings, earrings, and necklaces. Our method can generate realistic virtual try-on results as well as high-fidelity preservation of ornament identity.
    \item We propose an iterative scheme to estimate a pose-aware wearing mask to significantly improve the pose and scale alignments between ornaments and models. It also facilitates virtual try-on applications without the requirements of various additional inputs such as silhouette, semantic, and skeleton maps.
    \item By constraining attention to learning a mapping between reference and wearing masks, our method improves geometric feature preservation, especially for tiny items with complicated geometric structures, such as ornaments. 
\end{itemize} 




\section{Related Works}
\label{sec:formatting}
\noindent\textbf{Personalized image generation }
To address the challenge of text prompts not fully capturing user intent, personalized image generation, and editing have garnered significant attention from researchers. Inversion-based methods, such as Textual Inversion~\cite{TI} focus on optimizing a special prompt word to represent a target concept. Meanwhile, fine-tuning approaches~\cite{ruiz2023dreambooth,kumari2023multi,shi2024instantbooth,han2023svdiff} adjust pre-trained diffusion models using a small set of images of the target concept to create personalized models. Although these methods can generate high-fidelity images, they do not allow users to control the generation area for the target, and the additional optimization and fine-tuning required during inference can hinder large-scale applications. On the other hand, some zero-shot and training-free personalized image generation methods~\cite{wang2024instantid,ye2023ip-adapter}, can produce images of target concepts. However, they excel in stylized generation by representing high-level semantic features in the lack of preserving details and identity.
\begin{figure*}
    \centering
    \includegraphics[width=0.99\textwidth]{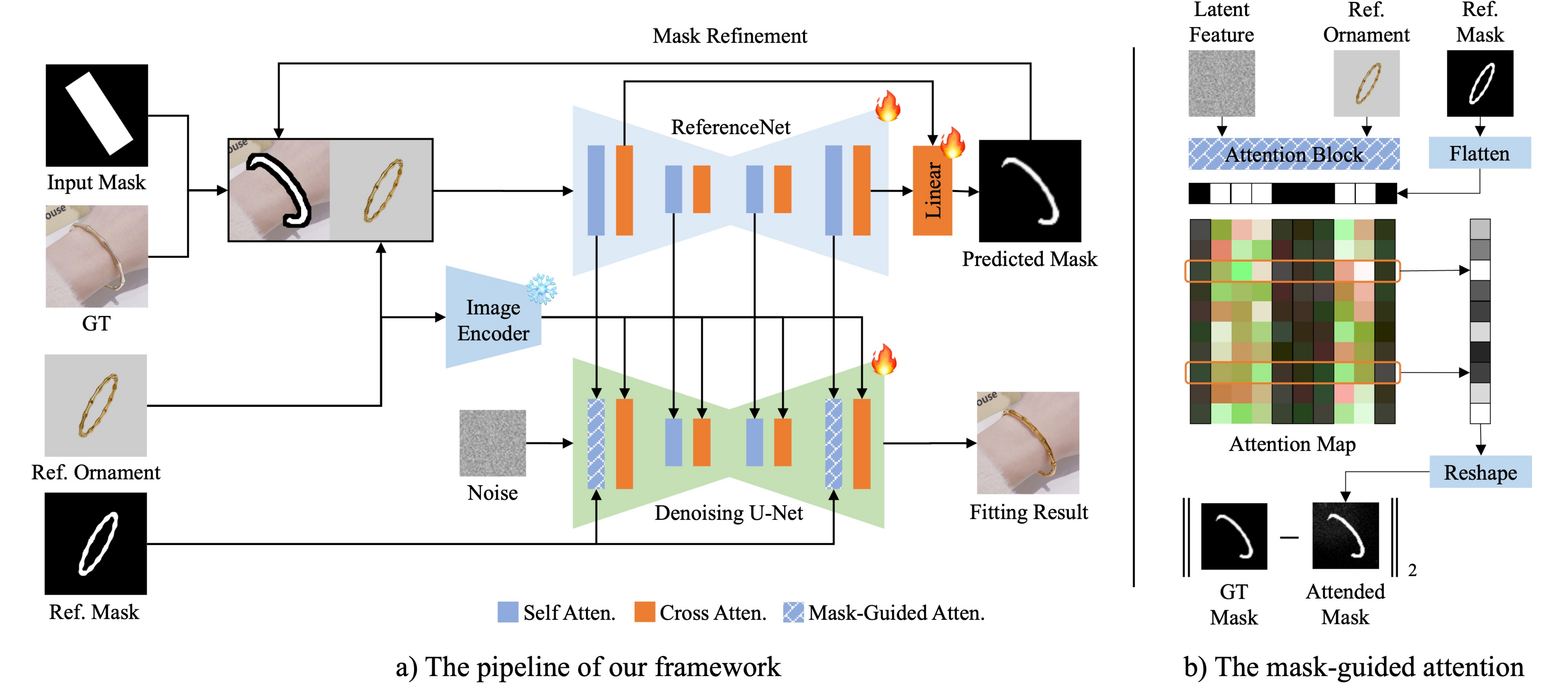}
    \caption{\textbf{The overview of our method.} a) In training, given reference ornament and model images and masks, our method concatenates ornament and masked model images as input to the ReferenceNet branch, which extracts features to predict wearing mask in an iterative way. The extracted features are also injected into the denoising U-Net to improve details generation. b) We enforce the attention layers to preserve structure details by formulating the layers to map the reference ornament mask to the ground truth wearing mask in an implicit way rather than directly imposing the mask onto attention maps.}
    \label{Structure}
\end{figure*}

\noindent\textbf{Local image editing }Our virtual ornament-wearing task closely resembles local image editing techniques. However, most previous local image editing methods relied on text prompts, such as Blended Diffusion~\cite{avrahami2022blended} and Blended Latent Diffusion~\cite{avrahami2023blended}, which employed multi-step semantic blending during denoising to produce harmonious images containing target semantic information within a defined mask. Inpainting Anything~\cite{yu2023inpaint} replaces any object in the input image with a target described by text prompts. However, for virtual ornament wearing, it is crucial to ensure alignment with the intricate details of the target. The traditional image composition pipeline~\cite{chen2019toward,cong2020dovenet} involves cutting and pasting foreground images onto background images, followed by the harmonization techniques. Recently, numerous diffusion-based methods~\cite{yang2023paint,song2023objectstitch,chen2024anydoor,winter2024objectdrop} have emerged in this field, significantly enhancing the quality and coherence of generated images. For instance, Paint by Example~\cite{yang2023paint} employs a CLIP image encoder~\cite{clip} to convert reference images into embeddings for guidance, generating objects that are semantically aligned with the reference image. ObjectStitch~\cite{song2023objectstitch} similarly utilizes CLIP to align text and images, guiding the generation of the diffusion model. ObjectDrop~\cite{winter2024objectdrop} first trains an object removal network, assembles a large dataset, and subsequently conducts object insertion training. AnyDoor~\cite{chen2024anydoor} leverages ControlNet and a DINO~\cite{caron2021emerging} encoder to extract detailed semantic information, improving its ability to maintain object identities. These methods focus on general object insertion into the background smoothly without pose alignment, which is normally required in virtual try-on tasks, especially for ornament wearing that needs precise pose alignment between the ornament and wrist, finger, or neck.

\noindent\textbf{Virtual try-on } Virtual try-on~\cite{zhu2023tryondiffusion,kim2024stableviton,xu2024ootdiffusion,yang2024texture,idm,zeng2024cat} takes a model image and an item image to generate an image of the model wearing the item. Early virtual try-on methods were based on Generative Adversarial Networks (GANs). Recently, with the significant success of diffusion models, researchers have explored their application in the field of virtual try-on. Most works focus on virtual try-on of garments, such as TryonDiffusion~\cite{zhu2023tryondiffusion}, OOTDiffusion~\cite{xu2024ootdiffusion}, and IDM-VTON~\cite{idm}. These methods utilize two parallel U-Nets for garment feature extraction, integrating them through self-attention and achieving impressive results. StableVITON~\cite{kim2024stableviton} introduced a zero-initialization cross-attention module to inject garment features into the denoising network. A few works~\cite{chen2024shoemodel,xue2024strictly} explore virtual try-on of shoes and earrings. In their settings, either the shoe pose is fixed to be aligned by the model or the earring has an almost vertical pose as it hangs down from the ear. Ornament virtual try-on requires pose alignment with different body parts at various poses and scales. Methods such as OOTDiffusion and IDM-VTON use additional inputs such as skeleton and semantic maps to guide wearing pose. In contrast, it's much more difficult to depict ornament wearing mask due to complicated tiny geometry structures and pose-related occlusion.



\section{Methodology}
We propose a zero-shot method for ornament virtual try-on with a reference ornament image, a target model image, and a coarse bounding box. The bounding box coarsely indicates the wearing location as ornament wearing is user-specific (e.g., a ring has its finger symbolism). We can generate realistic and high-fidelity fitting effects without additional inputs such as pose and semantic maps. The model comprises two vital components: 1) an iterative pose-aware wearing mask prediction and refinement module from the bounding box, which improves pose alignments between the ornament and the model; 2) a mask-guided attention module to improve identity and detail preservation. The framework of our method is illustrated in Fig. \ref{Structure}a. 

\subsection{Diffusion model and ReferenceNet}
\label{pre}
\noindent\textbf{Diffusion model} Our method is built upon the latent diffusion model (LDM) and ReferenceNet module, which has been widely adopted for condition generation and virtual try-on tasks. A typical LDM implementation~\cite{ldm} comprises an encoder-decoder module and a denoising network. The encoder embeds the input image into a low-dimension latent code to reduce computational overhead, which is diffused and then denoised by the denoising network to recover from a random noise. The denoised latent code is then decoded to generate an RGB image. The training process is formalized as follows:
\begin{equation}
    \mathcal{L}_1 = \mathbb{E}_{z_0,c,\epsilon,t}(||\epsilon-\epsilon_\theta(z_t,c,t)||_2^2)
\end{equation}
where $z_t$ represents the latent feature at time step $t$, which can be obtained through $z_t = \sqrt{\Bar{\alpha_t}}z_0+\sqrt{1-\Bar{\alpha_t}}\epsilon,\ \epsilon\in\mathcal{N}(0,I)$. $c$ is the condition embedding from a text prompt or reference image with a text or image encoder, and injected into the cross-attention layers to guide the generation. Our model adopts the widely used CLIP image encoder to extract features of the reference ornament image.

\noindent\textbf{ReferenceNet module} The module is widely used in virtual try-ons to improve detail and structure preservation. It is designed to be similar and parallelized with the denoising U-Net. The module extracts hierarchical latent features of the reference image which are injected into related layers in the denoising network. Specifically, latent features in the ReferenceNet are concatenated onto their counterparts in the denoising network for attention calculation. 


\subsection{Pose-aware Mask Refinement}
\label{maskgen}
We conducted several experiments to explore how the pose and scale impact the generative results, 
from which we have several key findings: 1) The diffusion-based model has the capability to fit ornaments to various model poses even without finetuning on ornament try-on datasets, which is attributed to the image prior from the pre-trained diffusion base model. 2) Poses and scales have significant influences on the fitting effects. In general, using an accurate wearing mask will significantly improve the pose alignment between ornaments and models even with large poses and scale variances. However, the wearing mask is not equivalent to the semantic mask, which is predicted from an existing image while wearing mask is hallucinated from two irrelevant ornament and model images. It's difficult or even infeasible to obtain accurate wearing mask in inference. In addition, ornaments usually show close-up views, which require a much more accurate wearing mask than the coarse silhouette mask in garment virtual try-on. 


\begin{figure*}
    \centering
    \includegraphics[width=\textwidth]{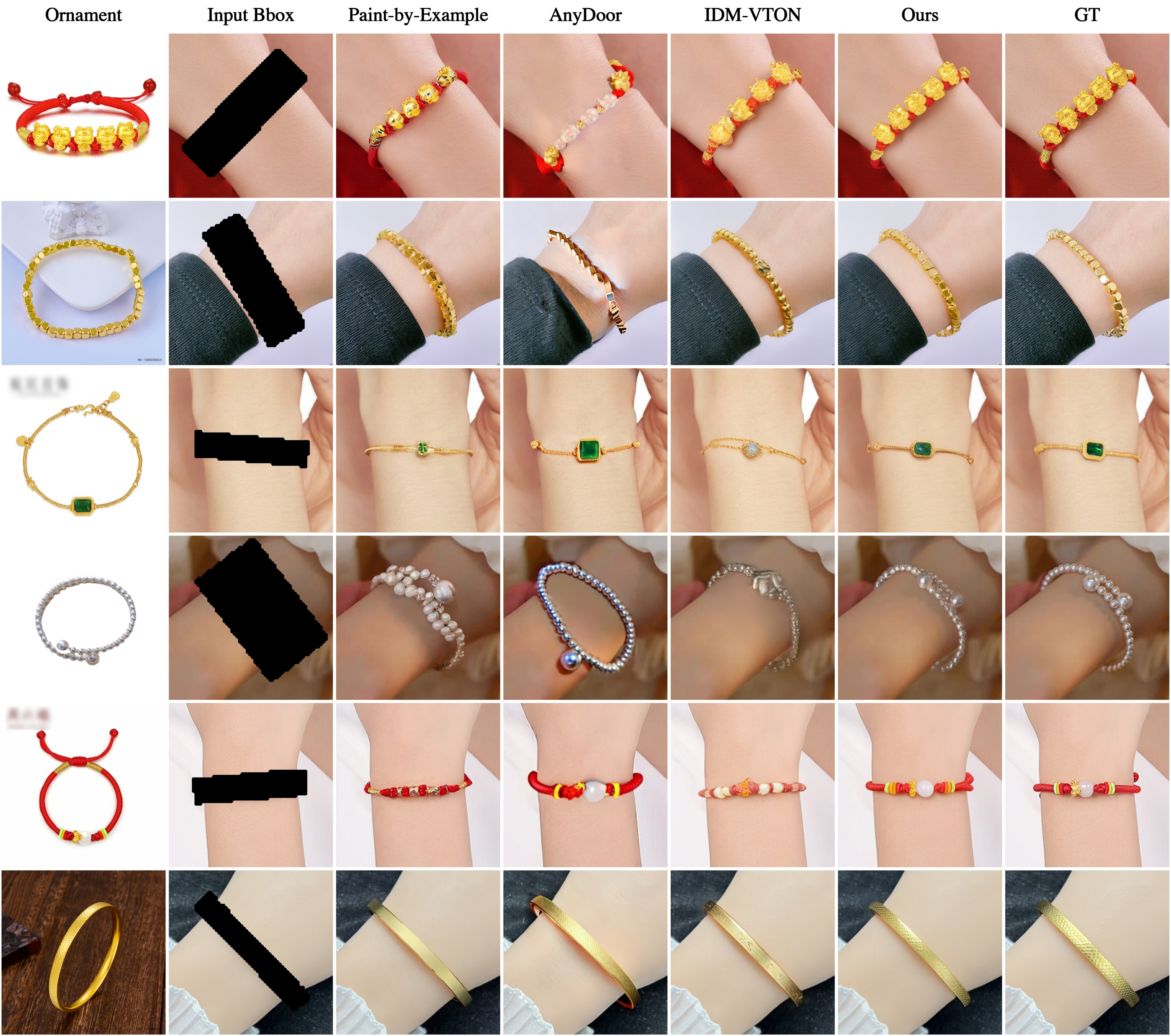}
    \caption{Visual comparison between previous methods and ours. No existing method could keep appearance and structure consistent, especially geometric details and numbers of components in ornaments. Our method preserves both details and identity and achieves high-quality and high-fidelity fitting results.}
    \label{test}
\end{figure*}

Previous works~\cite{hertzprompt, Tumanyan2023} have shown that intermediate results (e.g., latent features and attention maps) in early generative phases contain the semantic structure of the generated images. It might be possible to extract a wearing mask from these intermediate maps. However, extracted masks are too coarse to be used. To solve the problem, we proposed to estimate a more accurate wearing mask. We add an additional linear layer to predict the wearing mask from the intermediate maps. The predicted wearing mask is further used as input to guide the generation. The iterative refinement converges to an accurate pose-aware mask aligned with the model in the final generative image. Specifically, triplet images of an ornament, a model, and a coarse bounding box $M_{b}$ are fed into the ReferenceNet. The ornament latent features $f_o^t$ and the model latent features $f_m^t$ are injected into the denoising counterpart similar to most networks in garment virtual try-ons. The latent features are further concatenated and linearly projected to predict the wearing mask $\hat{M}_p^t$. The predicted mask and the bounding box are blended as a new wearing mask input, which is updated as follows:  
\begin{equation}
     \hat{M}_p^{t-1} = \alpha_t\odot\hat{M}_p^{t-1} + (1-\alpha_t)\odot M_{b}
\end{equation}
\begin{equation}
     \hat{M}_p^t = \text{MLP}([f_m^t \odot \hat{M}_p^{t-1}, f_o^t])\\
\end{equation}
where $\alpha_t \in [0, 1]$ is a hyperparameter in terms of training step $t$. In the early training, the predicted wearing mask is coarse, and $\alpha_t$ is set to be small. As the mask gets more accurate in the late stage, $\alpha_t$ approximates to 1.0.   

As mask prediction and image generation are entangled, we employ an ornament try-on dataset with wearing masks to regularize mask prediction with a $L_2$ loss:
\begin{equation}
    \mathcal{L}_2 = ||\hat{M}_p^T - M_o^{gt}||_2^2
\end{equation}
where $M_o^{gt}$ is the ground-truth wearing mask. The regularization is important to prevent the dual degeneration of both results due to mutual dependence. In inference, only a bounding box is required to indicate the user-specific wearing location.  

\subsection{Mask-guided Attention}
\label{aaa}
\label{train}
The precise wearing mask improves the alignment between ornaments and models with various poses and scales. It also has a positive effect on detail generation. However, most ornaments comprise complicated tiny geometric components such as repeated shapes and/or ring structures in pearl necklaces and beaded bracelets. Our early attempts found that the model had difficulty preserving the topology and/or the number of components, especially in repeated geometric patterns. To take a few examples, it may ignore small parts interleaved with other large components, or fill the hole of a ring structure. We suspect that existing generative networks can capture appearance and spatial details rather than geometric structures, as geometric shapes require hard constraints on local primitive structures of edges and contours.  

Attention maps retain shape details with affinities between the spatial features~\cite{hertzprompt}. A possible solution is to impose a geometric structure constraint in attention maps. The semantic segmentation mask contains rich geometric structure information, but the mask is difficult to extract due to massive tiny complicated sub-components in ornaments. As the binary mask is also full of geometric structures of edges and contours and easy to obtain (e.g., with SAM~\cite{kirillov2023segment}), we propose to employ the reference ornament mask to inject geometric structure into the generation.    

However, directly blending attention maps with the mask may mask out too much information to degrade generated results. We introduce an indirect way to restrict geometric structure changes of ornaments in reference and generated images. Specifically, we obtain attention maps $\{M_a^i\ \in \mathbb{R}^{d_i \times d_i}\}_1^N$ of latent features and ornament embeddings from various layers in denoising U-Net, where $N$ is the number of extracted attention maps and $d_i$ is the dimension of $i$-th attention map. The ornament mask $M_o$ in the reference image is down-sampled and flattened as one-dimensional masks $\{M_o^i\ \in \mathbb{R}^{d_i}\}_1^N$. We then apply $M_o^i$ to mask out the attention map $M_a^i$ along one dimension and margin it along the other dimension. The result is then reshaped and up-sampled to $\Tilde{M}_o^i$ as the same dimension of $M_o$. All result masks $\{\Tilde{M}_o^i\}$ are then averaged as the final mask $\Tilde{M}_o$, which can be formulated as:
\begin{equation}
        M_o^i =\textbf{T}_\text{flatten}^{\sqrt{d_i} \times \sqrt{d_i} \rightarrow d_i} \circ \textbf{T}_\text{downsampling}^{d_0 \times d_0 \rightarrow \sqrt{d_i} \times \sqrt{d_i}} \circ M_o
\end{equation}
\begin{equation}
        \tilde{M}_o^i = M_a^i \odot \underbrace{[{M_o^i}^T, ..., {M_o^i}^T]}_{d_i}
\end{equation}
\begin{equation}
        \tilde{M}_o^i =\textbf{T}_\text{upsampling}^{\sqrt{d_i} \times \sqrt{d_i} \rightarrow d_0 \times d_0} \circ \textbf{T}_\text{reshape}^{d_i \rightarrow \sqrt{d_i} \times \sqrt{d_i}}  \circ \sum_{c} \tilde{M}^i_{o}[r][c]
\end{equation}
\begin{equation}
        \tilde{M}_o = \frac{1}{N}\sum_i^N \tilde{M}_o^i 
\end{equation}
where $\textbf{T}_\textbf{ops}^{\textbf{d1} \rightarrow \textbf{d2}}$ is the operator with $\textbf{ops}$ as operation type and $\textbf{d1} \rightarrow \textbf{d2}$ as dimension mapping from \textbf{d1} to \textbf{d2}. Sequential operators are defined to execute from right to left. The masking operation enforces latent features to attend to ornament regions in the reference image, while the margining operation diffuses ornament features to the wearing region in the generated image. The reference mask $M_o$ is mapped to the wearing masks $\tilde{M}_o$ via the attention map. Inversely, in order to enable the attention map to learn the mapping, we require the transformed wearing masks $\tilde{M}_o$ to be consistent with the ground truth wearing ornament mask $M_o^{gt}$ with an $L_2$ loss as below:      
\begin{equation}
    \mathcal{L}_3 = ||\Tilde{M}_o - M_o^{gt}||_2^2
\end{equation}
The process is shown in Fig.\ref{Structure}b. Down-sampling and up-sampling operations are not displayed for concise illustration.
\subsection{Training}
\label{train}
\noindent\textbf{Dataset} 
Inspired by the common practice in garment virtual try-ons, we collect image pairs of ornaments and models wearing the ornaments. We mask out the ornament in the model image to obtain the target image and ground-truth wearing mask. The reference ornament image, the model image with masked-out ornament, and the original model image are combined as a training triplet image. We also label the masks in ornament images as reference masks. Our dataset does not require pose alignment between the ornament and the model, which is easy to collect, and also prevents the model from learning a simple copy-and-paste strategy. In total, we collect about 64k image triplets, roughly evenly distributed over four categories of bracelets, rings, earrings, and necklaces. Each image triplet also contains a reference mask and a wearing mask of the ornament.

\noindent\textbf{Training loss} Our training loss comprises the aforementioned three items :
\begin{equation}
\begin{aligned}
     \mathcal{L}_{total} =& \mathcal{L}_1  + \lambda_1 \mathcal{L}_2 +  \lambda_2\mathcal{L}_3
\end{aligned}
\end{equation}
where $\lambda_1$ and $\lambda_2$  are loss weights. The two weights decay as the training step increases, which forces the model to learn the wearing mask in the early stages. As the mask becomes accurate, the model focuses on the generation of appearance details. The scheme follows common observation~\cite{tewel2024training} that image layout and structure are sketched in the early stage and details are generated in the late stage. 

\section{Experiments}
\begin{figure*}
    \centering
    \includegraphics[width=\textwidth]{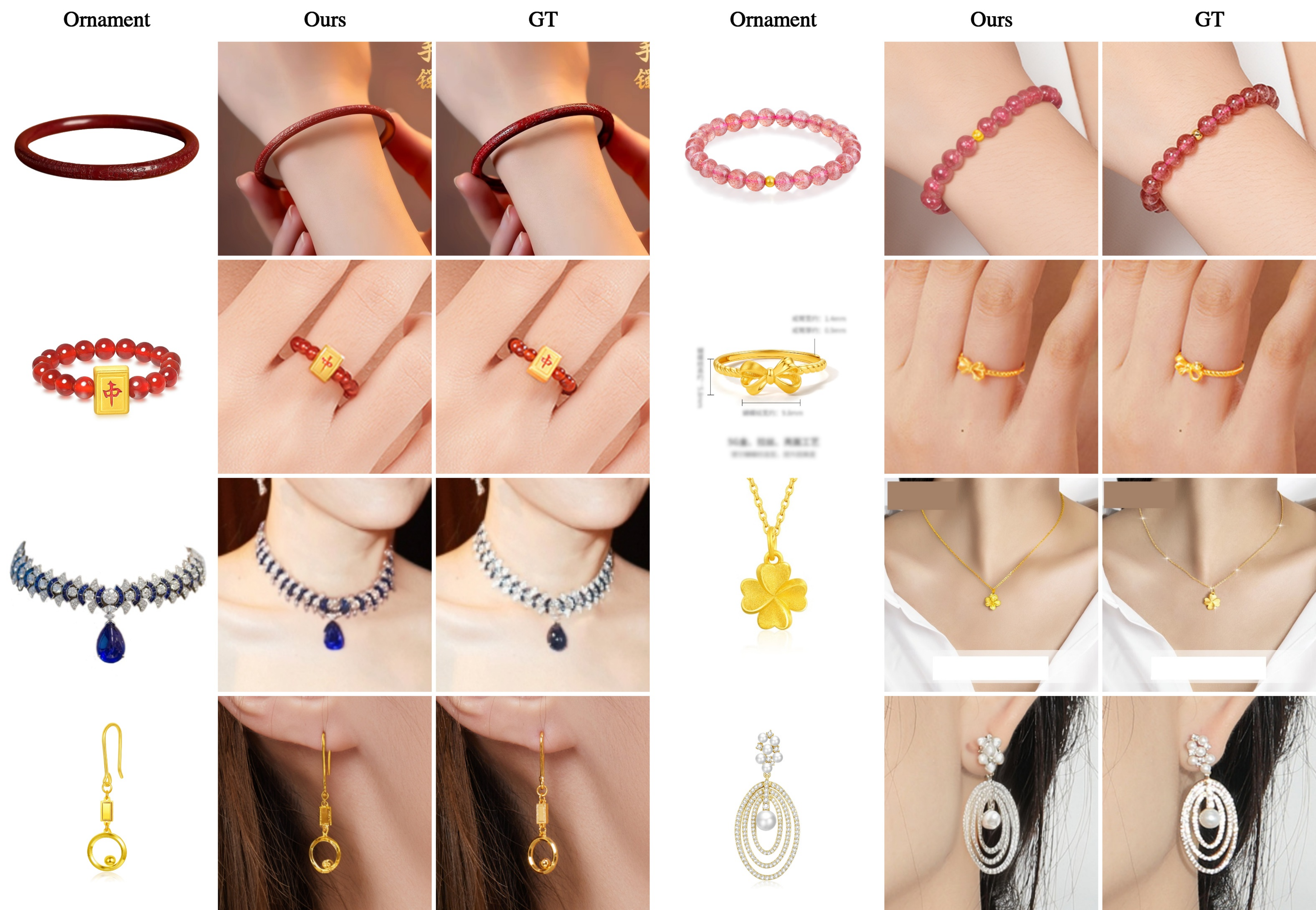}
    \caption{Virtual try-on results on other categories including bracelets, rings, necklaces, and earrings.}
    \label{other}
\end{figure*}
\begin{table*}[]
\centering
\caption{Quantitative evaluations between our and other methods}
\label{wild_table}
\begin{tabular}{c|cccc|cc}
\hline
& \multicolumn{4}{c|}{Compared against ground truth } & \multicolumn{2}{c}{Compared against reference ornament}\\\hline
Method           & FID$\downarrow$& LPIPS$\downarrow$&CLIP Score$\uparrow$ & DINO Score$\uparrow$ &CLIP Score$\uparrow$ & DINO Score$\uparrow$\\ \hline
Paint-by-Example~\cite{yang2023paint} & 23.49&0.0789& 85.6    &   64.8   &   57.2       & 35.4  \\
AnyDoor~\cite{chen2024anydoor}          & 28.28 & 0.1029 &85.1    &    67.2  &     54.8     & 35.9 \\
IDM-VTON~\cite{idm}         & 22.99 & 0.0709 & 85.9   &  65.0   & 55.9      & 35.2 \\ 
Ours    & \textbf{19.00} &\textbf{0.0593}&\textbf{88.7} & \textbf{74.5}   &   \textbf{57.3}   &  \textbf{38.7}    \\ \hline
Ground Truth      & - &-&   -       &  -     & 59.0  & 43.3  \\ \hline
\end{tabular}
\end{table*}
\subsection{Implement details}
Our model adopts the Stable Diffusion V1.5 as the network backbone. The ornament region is cropped and resized to $512\times512$, and Adam~\cite{kingma2014adam} optimizer is chosen with an initial learning rate of $1e^{-5}$. We employ a simple linear decay for the $\alpha(t)$ and select the self-attention maps with the highest resolution from the encoder and decoder for masked-guided attention. We found these simple settings are enough to obtain compelling results. We follow a similar scheme as AnyDoor~\cite{chen2024anydoor} in handling inputs and composing final results. Specifically, the ground truth mask is resized to be square, and the cropped image is scaled by a factor of 1.5. The generated result is pasted back to the original masked region to compose the final result. Our model takes about 10 hours to train on 8 A100 GPUs with 10 epochs.  

For quantitative comparison, we adopt FID~\cite{heusel2017gans} and LPIPS~\cite{zhang2018unreasonable} to evaluate image quality, while the CLIP image similarity score and DINO-based feature similarity score are used to measure the identity consistency of the ornaments. All results are calculated and averaged on a test image set split from our dataset. 
\subsection{Comparisons}
As we are the first to focus on ornament virtual try-ons, we select several works that are most related to ours in the broad field of image edits. These works include Paint-by-Example(CVPR'23)~\cite{yang2023paint}, AnyDoor(CVPR'24)~\cite{chen2024anydoor}, and IDM-VTON(ECCV'24)~\cite{idm}. The first two works are designed to insert reference objects into target images, while the latter is dedicated to garment virtual try-ons. Similar to ours, these methods require a reference image of the item and a target image as well as masks to define local edit regions. All methods are trained or fine-tuned with our dataset. Limited to the page length requirement, we take the bracelet category as an example to illustrate all visual results. Please refer to the Appendix for more results of other ornament categories. 

\noindent\textbf{Qualitative results} Fig. \ref{test} qualitatively compares fitting results on ornaments of various structures and poses. Paint-by-Example could hardly preserve geometric structures and appearance in most cases. AnyDoor struggles to preserve the scales of the whole structure and/or major parts. IDM-VTON could preserve the scale to a certain extent, but it has problems maintaining structure layouts, especially for complicated ornaments with multiple parts. None of the previous methods could hold the number of sub-parts in ornaments or recover repeated geometric patterns. Our method has the most visual appearance and structure similarities to both reference ornaments and ground truth images, which indicates its ability to preserve both appearance and local and global structures, and tiny surface geometric patterns (the last row). Surprisingly, our results seem to be biased towards reference ornaments with less specular reflections than ground truth. It's partially because model images do not have enough hints of environmental illumination that our model has difficulty in learning the exact light effect as the ground truth. 

\noindent\textbf{Quantitative results} Table \ref{wild_table} illustrates a quantitative comparison between ours and previous methods. The two ornaments in reference and ground-truth images are usually captured with different conditions such as views and illumination. Therefore, we compare the results generated by all methods against both the reference ornaments and the ground truth, and calculate the corresponding consistency metrics. Our approach achieves the best results in all metrics, demonstrating its capability to generate more realistic and high-fidelity virtual results.
\subsection{Ablation Study}
\begin{figure}
    \centering
    \includegraphics[width=\linewidth]{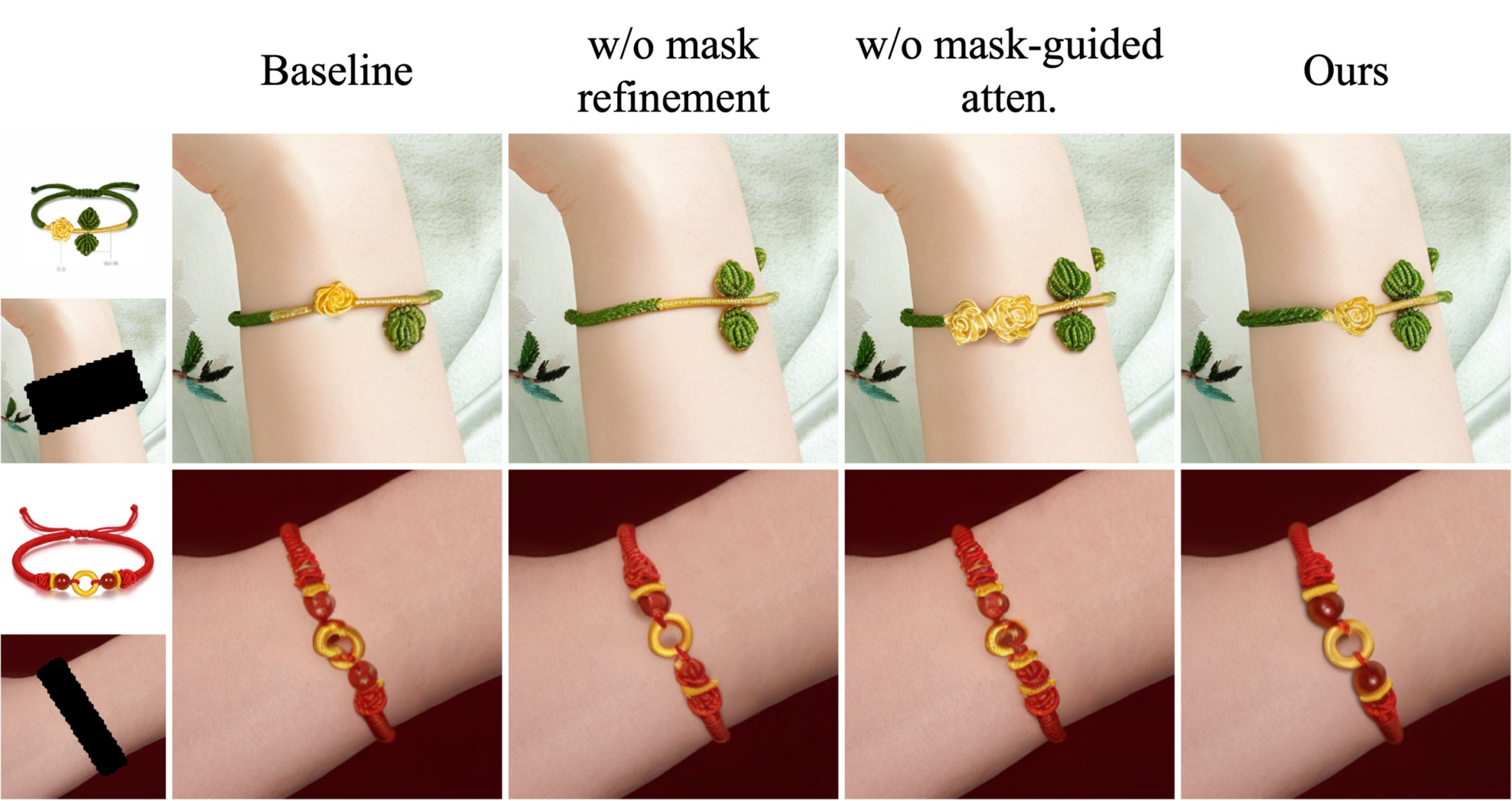}
    \caption{The visual comparisons of our models with different module configurations. The full model archives the best results with the proposed two modules.}
    \label{ab_attn}
\end{figure}

\begin{table}[]
\centering
\caption{Quantitatively comparisons of our models with different module configurations.}
\label{attn_abl}
\begin{tabular}{c|cc}
\hline
Method        & CLIP Score$\uparrow$ & DINO Score$\uparrow$ \\ \hline
Baseline  &   86.9    &      71.9      \\
w/o mask refinement &  88.5  &  73.3          \\
w/o mask-guided atten.  &     88.0 &    73.9     \\
Ours    &  \textbf{88.7}   & \textbf{74.5}   \\ \hline
\end{tabular}
\end{table}
We conduct a comprehensive ablation study to evaluate the effectiveness of our proposed components. The experiments are designed by adding a component from the basic models. The baseline is adapted from ReferenceNet and Stable Diffusion. The results are evaluated from both qualitative and quantitative aspects. Fig. \ref{ab_attn} shows the visual comparisons with various component configurations. The basic model has noticeable defects in details and geometric structures. If we do not integrate the mask prediction, the results lack appearance details and specular lights. It may also lose structure consistency to some extent (e.g., flow structure missing in the first ornament). Without mask-guided attention, both the local and global structures are destroyed in form of adding or missing components as well as changing scales. On the other hand, the full model preserves both appearance and geometric details as well as global structures. The quantitative results in Table \ref{attn_abl} also indicate the importance of the two proposed modules to improve the final virtual try-on results. More results are in the Appendix.
\begin{figure}
    \centering
    \includegraphics[width=\linewidth]{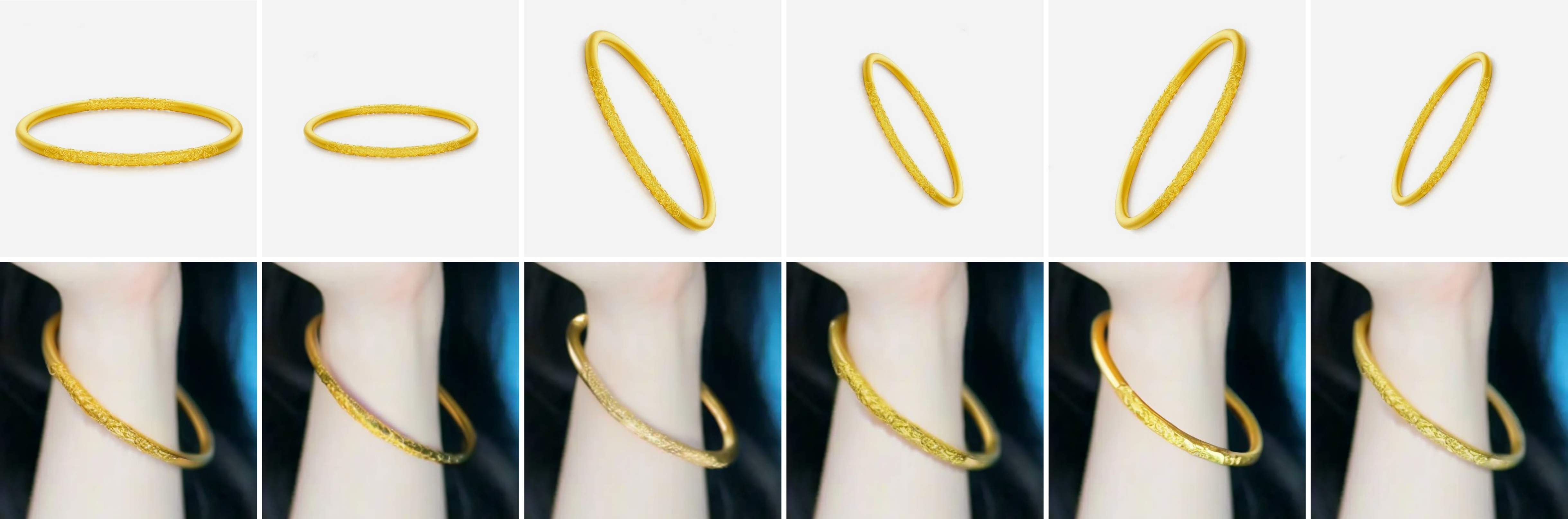}
    \caption{Our model is robust to achieve consistent results with different poses and scales.}
    \label{scale}
\end{figure}
\subsection{More Results}
We use our method to wear various types of ornaments including bracelets, necklaces, earrings, and rings. Figure \ref{other} lists the results. Other configurations including a model wearing different ornaments and different models wearing the same ornaments are also illustrated in Figure \ref{first} (the last row). All results demonstrate that our method can handle various ornament structures of local and global rigid and non-rigid components. More results are in the Appendix.

To evaluate the robustness of our model under conditions of different poses and scales. We also conduct experiences by randomly rotating and scaling the reference ornament, which is then used to wear on the same model. As Fig. \ref{scale} shows, the results are consistent in details and geometric structures with different configurations. The experiment results show our model is robust with large pose differences between ornaments and models. 
\section{Conclusion}
We propose the virtual try-on ornament task for the first time. To tackle the more challenging problems of intricate geometric structures in ornaments, we devise two modules of mask prediction and mask-guided attention to obtain accurate wearing masks and impose geometric structures, which preserve both appearance details and geometric structures to achieve identity consistency. 
Currently, our method is biased toward reference images rather than ground truth images, which lack specular reflections to a certain extent. Inspired by the work~\cite{zeng2024dilightnet}, we would like to add more fine-grained lighting control in our future work. Besides, inaccurate control over wearing orientations (e.g., rotated along a wrist) leading to featured components being hidden behind the wrist occasionally. Secondary masks and local feature injection into the diffusion process may fix the problem.

\section*{Acknowledge}
This research is supported by Sensetime, the National Key Research and Development Program of China (2023YFB3107401), the National Natural Science Foundation of China (T2341003, 62376210, 62161 160337, 62132011, U21B2018, U24B20185, 62206217), the Shaanxi Province Key Industry Innovation Program (2023-ZDLGY-38). Special thanks to Jessie Geng for the coordination of computing resources.

{
    \small
    \bibliographystyle{ieeenat_fullname}
    \bibliography{main}
}

\clearpage
\setcounter{page}{1}
\maketitlesupplementary

\section{Additional Experiments}
\label{sec:rationale}

\subsection{Comparison on Necklaces, Earrings, and Rings}
\begin{figure*}
    \centering
    \includegraphics[width=0.95\textwidth]{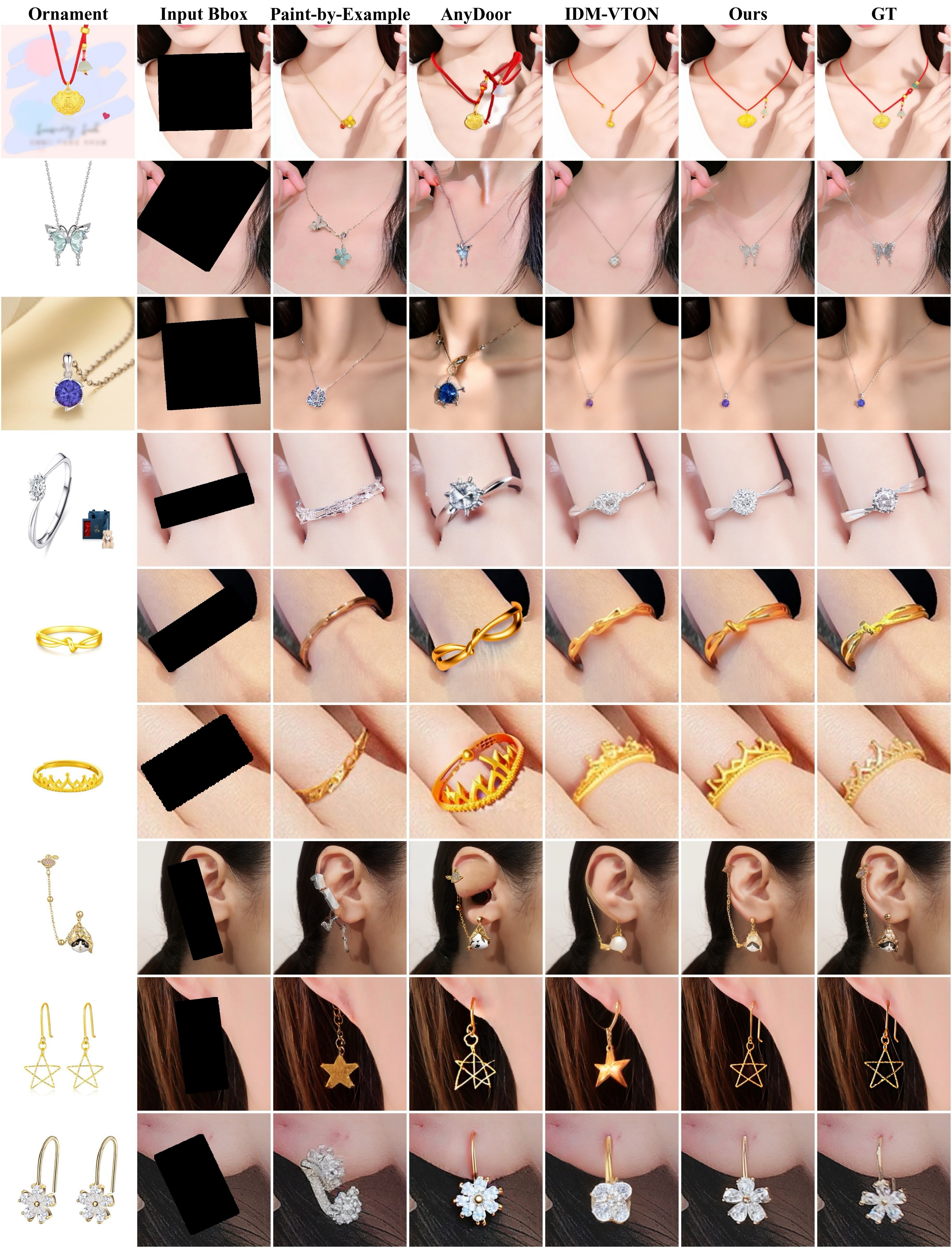}
    \caption{Visual comparison on necklaces, rings, and earrings.}
    \label{com_other}
\end{figure*}

In addition to the comparison with previous methods using the bracelet dataset, as discussed in the main text, we extended the evaluation to other categories of ornaments. The results are shown in Fig. \ref{com_other}. Consistent with the qualitative comparisons in the main text, Paint-by-Example \cite{yang2023paint} faces challenges in preserving geometric or ID information, retaining only partial semantic features. AnyDoor \cite{chen2024anydoor} has difficulty capturing correct wearing patterns and natural region inpainting. IDM-VTON \cite{idm} shows some improvement over other methods but still struggles with maintaining fine details and spatial relationships.

In contrast, our method demonstrates superior performance across all categories. Necklaces, due to their small wearing area and the challenge of invisible chain parts, are the most difficult among all categories. Nevertheless, our method achieves good results. Notably, in the first row of examples, where some structures are not visible in the reference image, our method faithfully preserves the reference ornament's ID information. 
Rings share a similar wearing pattern to bracelets but include more subtle structural details. As seen in the 5th and 6th rows, our method generates high-fidelity try-on images that preserve these fine details. 
For earrings, the results in the 8th row demonstrate that our method effectively handles fine linear decorations.

\subsection{Additional Results}
\begin{figure*}
    \centering
    \includegraphics[width=0.95\textwidth]{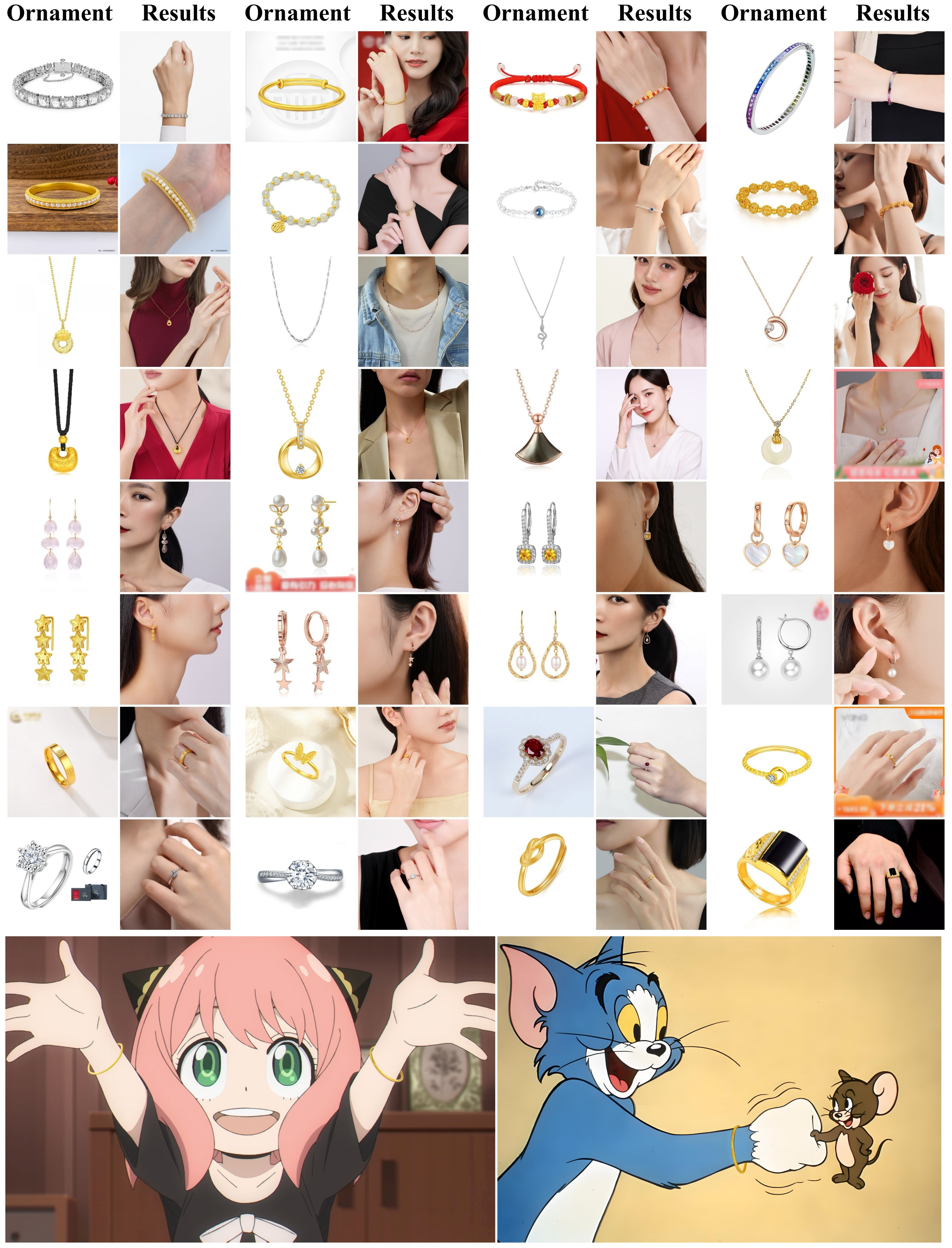}
    \caption{More results.}
    \label{more}
\end{figure*}

We conducted further experiments across all categories, and Fig. \ref{more} presents the results of our virtual try-on. Our method demonstrates stable, high-fidelity virtual try-on for ornaments. Additionally, we explored cross-domain ornament virtual try-on and found that our method is capable of virtual try-on for certain animated characters. This showcases the robustness of our approach and indicates that, with sufficient data, our method can be extended to images from other domains.

\subsection{Additional Ablation Study}
\begin{figure}[!ht]
    \centering
    \includegraphics[width=\linewidth]{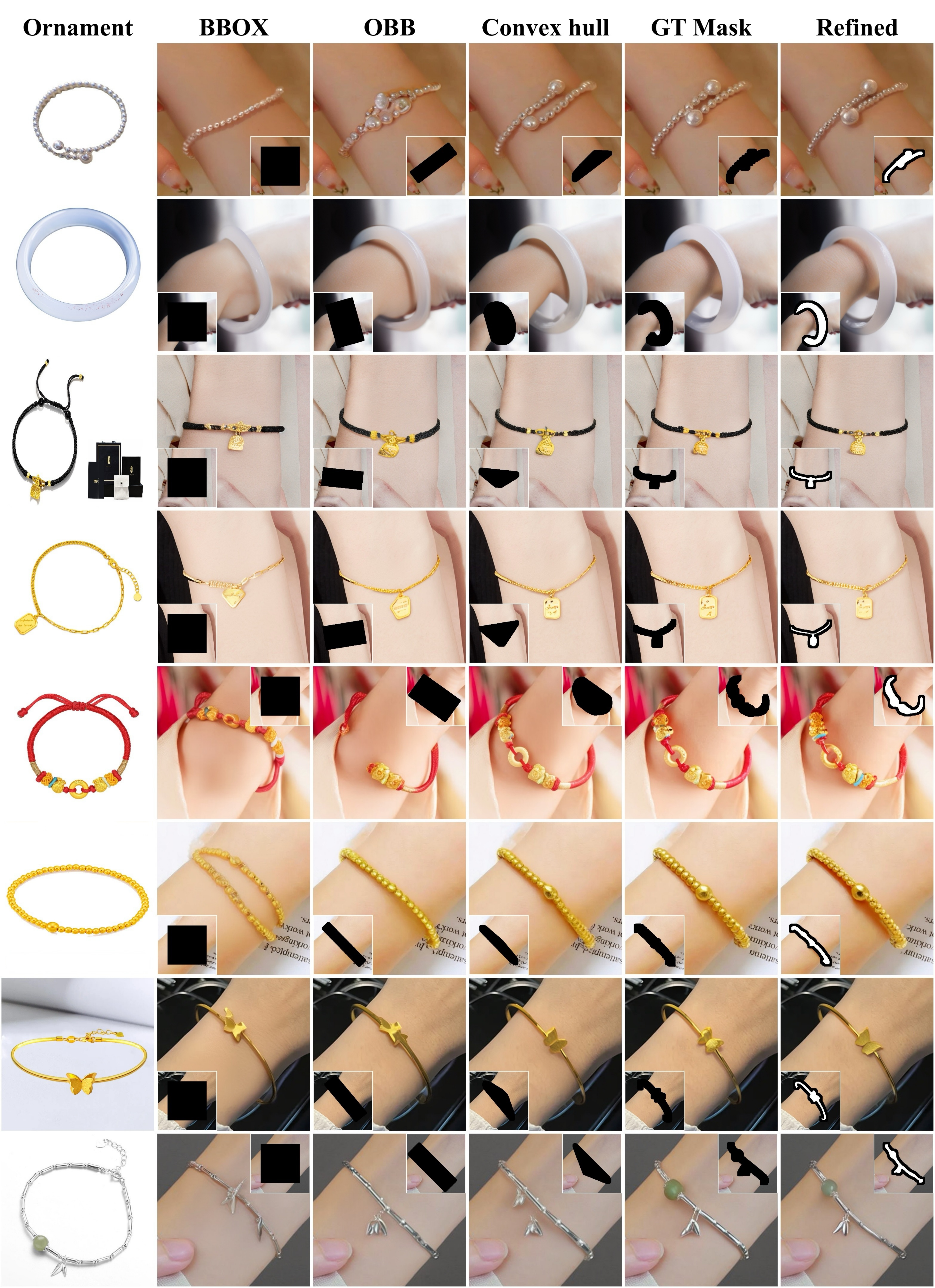}
    \caption{Experiments to validate our motivation.}
    \label{mask_abl}
\end{figure}

\noindent\textbf{Experiments to Validate Our Motivation.} To validate the motivation for mask refinement, we conducted experiments on different masks using the baseline based on ReferenceNet and Stable Diffusion. The results are shown in Fig. \ref{mask_abl}. We trained the model with various input masks, including bounding boxes, oriented bounding boxes (OBB), convex hulls, and ground truth masks.
The results show that as mask refinement increases, the fidelity and ID consistency of the generated images improve. For instance, using a bounding box as the mask increases the probability of errors in the wearing pattern. The OBB, which adds limited pose information, shows some improvement but still fails to produce satisfactory results. The convex hull adds extra shape information, and some geometric structures, which earlier methods struggled to preserve, are retained in the output. The ground truth mask yields the best results, providing precise shape and location information. These results highlight the importance of pose and accurate masks for generation quality, which is a key idea in our work. However, as mentioned in Section \ref{maskgen} of the main text, a fine-grained wearing mask cannot be obtained during the inference process. 
Our method uses only the bounding box as input to predict the wearing mask and then refines the original mask. The results demonstrate that our method achieves performance close to that of the ground truth mask.

\subsection{Results on Garments Virtual Try-on}
\begin{figure}[ht]
\centering
\includegraphics[width=\linewidth]{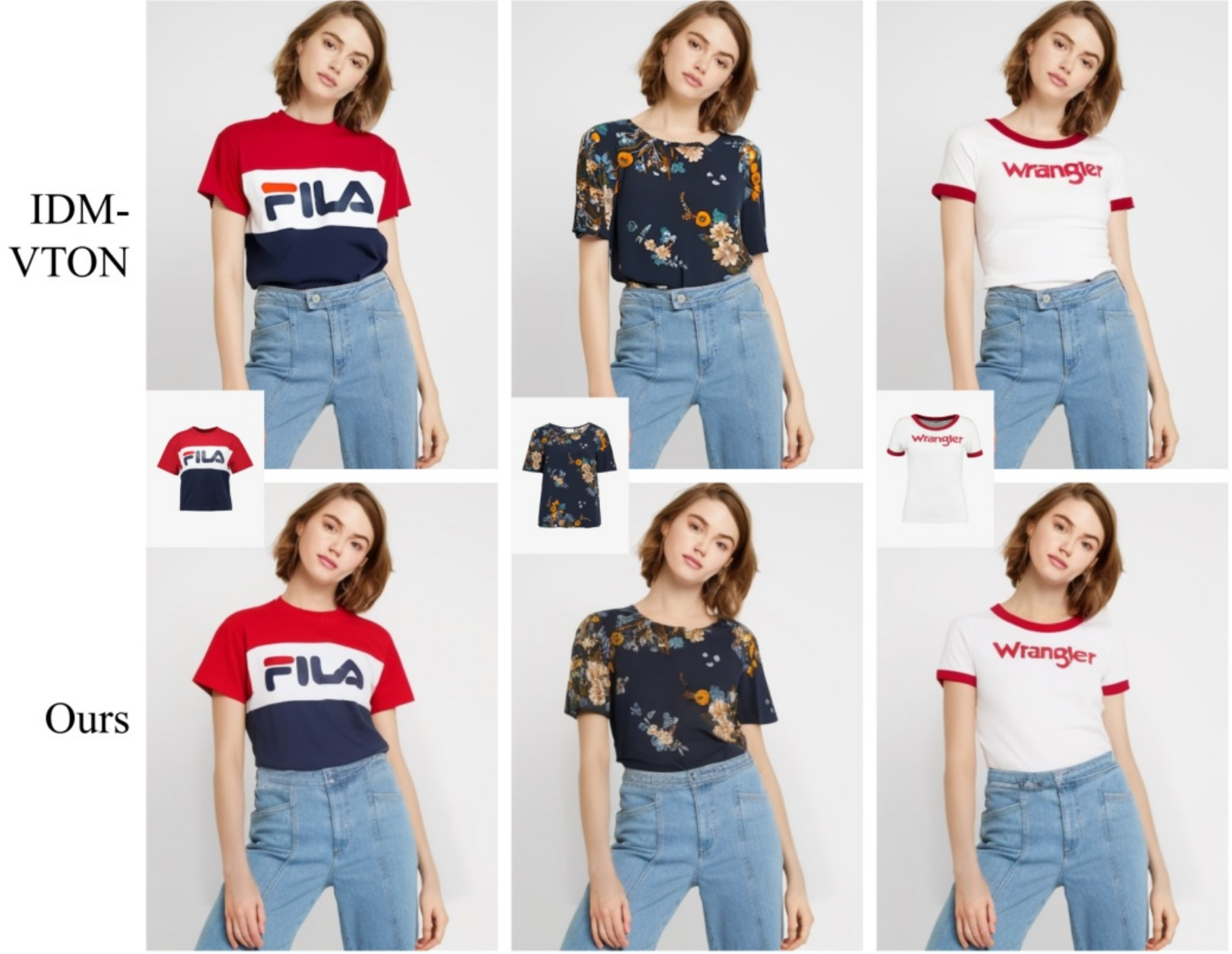}
\caption{Visual comparison on garments virtual try-on. }
\label{garments}
\end{figure}

Our task extends the field of garments virtual try-on. As analyzed in the introduction, the proposed ornaments virtual try-on presents more significant challenges compared to garments virtual try-on. Therefore, we have developed a series of customized improvement strategies. To demonstrate the performance of our method in garments virtual try-on, we conducted qualitative experiments on the VITON-HD dataset, and the comparative results with IDM-VTON are shown in Fig.\ref{garments}.

It should be noted that our method is not specifically designed for garments virtual try-on, and thus many inputs for garments virtual try-on are not considered. We only utilize the model image, garments image, and mask. Despite this limitation, our method achieves comparable results to state-of-the-art (SOTA) methods. Given that garments virtual try-on has been extensively studied and previous methods have achieved remarkable results, our method faces certain limitations in further enhancing garments try-on performance.

Through multiple experimental validations presented in this paper, we can conclude that our method exhibits excellent versatility. It maintains the performance of garments virtual try-on while achieving outstanding results in ornaments virtual try-on. This characteristic makes our method highly promising for a wide range of applications in the field of virtual try-on.



\end{document}